# An Intelligent Fault Diagnosis Method for General Aviation Aircraft Based on Multi-Fidelity Digital Twin and FMEA Knowledge Enhancement

Zhihuan Wei [a], Yang Hu [a,*], Xinhang Chen [a], Yiming Zhang [b], Jie Liu [c], Wei Wang [d]

[a] *Hangzhou International Innovation Institute, Beihang University, Hangzhou, China*

[b] *School of Mechanical Engineering, Zhejiang University, Hangzhou, China*

[c] *School of Reliability and Systems Engineering, Beihang University, Beijing, China*

[d] *Department of Mechanical Engineering, City University of Hong Kong, Hong Kong, China*

Corresponding author. E-mail: yang_hu@buaa.edu.cn

**Abstract**

Fault diagnosis of general aviation aircraft faces fundamental challenges including scarce real fault data, diverse fault types, and weak fault signatures. This paper proposes an intelligent fault diagnosis framework based on multi-fidelity digital twin, integrating four core modules: high-fidelity flight dynamics simulation, FMEA-driven fault injection mechanism, multi-fidelity residual feature extraction, and large language model (LLM)-enhanced interpretable diagnostic report generation. First, a digital twin model is constructed based on the JSBSim six-degree-of-freedom (6-DoF) flight dynamics engine, generating 23-channel engine health monitoring data through semi-empirical sensor synthesis equations. A three-layer fault injection engine based on failure mode and effects analysis (FMEA) is designed to model the physical causal propagation of 19 types of engine faults. Second, a multi-fidelity residual computation framework comprising paired-mirror residuals and GRU surrogate prediction residuals is proposed: the high-fidelity path obtains clean fault deviation signals using nominal mirror trajectories with identical initial conditions, while the low-fidelity path achieves online real-time residual computation through a multi-step prediction GRU surrogate model. A 1D-CNN classifier is then employed for end-to-end diagnosis of 20 fault classes. Finally, an LLM diagnostic report engine enhanced with FMEA knowledge is designed, which fuses classification results, residual evidence, and domain causal knowledge to generate interpretable natural language diagnostic reports. Experimental results demonstrate that the paired-mirror residual scheme achieves a Macro-F1 of 96.2% on the 20-class fault classification task, while the GRU surrogate scheme achieves 4.3x inference acceleration at a cost of only 0.6% performance degradation. Comprehensive comparison experiments across 24 schemes reveal that the contribution of residual feature quality to diagnostic performance is approximately 5 times that of classifier architecture selection, establishing the "residual quality first" system design principle.

# 1. Introduction

General aviation aircraft play an irreplaceable role in flight training, short-haul transportation, agricultural and forestry operations, and emergency rescue, with their operational reliability directly related to flight safety [1-3]. Light general aviation aircraft, represented by the Cessna 172P, widely employ piston-powered propulsion systems. Their engines feature compact mechanical structures with numerous fault types spanning multiple subsystems including cooling, lubrication, intake, compression, and electrical systems [4,5]. Moreover, many faults exhibit weak sensor responses in their early stages, making them difficult to reliably detect using conventional threshold-based monitoring methods [6,7]. Therefore, developing data-driven intelligent fault diagnosis methods for general aviation aircraft holds significant engineering value [8,9].

However, the field of aero-engine fault diagnosis faces a fundamental data bottleneck: real fault samples are extremely scarce [10]. Most fault types (e.g., cylinder cracks, baffle structural damage) have extremely low actual occurrence frequencies, and accumulating a sufficient number of labeled samples often requires years or even longer operational periods [11,12]. In this context, digital twin technology provides a breakthrough path for fault diagnosis—generating simulation data covering various fault conditions on demand through high-fidelity physics-based simulation models, fundamentally alleviating the data scarcity problem [13,14].

Existing digital twin-based fault diagnosis research has primarily focused on rotating machinery, bearings, and gearboxes [15-18], while research in the aero-engine domain remains in its infancy [19,20]. Most existing works adopt simplified fault modeling approaches (e.g., directly superimposing bias or noise on sensor signals) without systematic modeling of fault physical propagation mechanisms [21]. Furthermore, as a core component of model-based fault diagnosis, the impact of different residual feature extraction paths on diagnostic performance has not yet been systematically and quantitatively evaluated [22].

Regarding interpretability, although deep learning classifiers can achieve high-accuracy fault identification, their "black-box" nature limits the credibility and acceptability of diagnostic conclusions in aviation safety-critical domains [23]. In recent years, the application of large language models (LLMs) in industrial intelligent maintenance has attracted extensive attention [24]. However, deeply integrating LLMs with structured domain knowledge (e.g., FMEA causal chains) to generate physically interpretable diagnostic reports remains an open research problem [25].

To address the aforementioned challenges, this paper proposes an intelligent fault diagnosis method for general aviation aircraft based on multi-fidelity digital twin and FMEA knowledge enhancement. The main contributions include:

(1) A digital twin model is constructed based on the JSBSim 6-DoF flight dynamics engine, integrating semi-empirical sensor synthesis equations and an FMEA-driven three-layer fault injection engine, achieving physical causal propagation modeling of 19 types of engine faults spanning five major subsystems: cooling, lubrication, intake, compression, and electrical.

(2) A multi-fidelity residual computation framework comprising paired-mirror residuals and GRU surrogate prediction residuals is proposed. The high-fidelity path obtains clean fault signals with a "zero-bias" baseline through nominal mirror trajectories with identical initial conditions, while the low-fidelity path achieves online real-time residual computation through a multi-step prediction GRU surrogate model. Both paths share a unified classifier interface, supporting flexible accuracy-speed trade-offs.

(3) Through comprehensive comparison experiments across 24 schemes (4 feature modes × 6 classifiers), it is quantitatively revealed that the contribution of residual feature quality to diagnostic performance is approximately 5 times that of classifier architecture selection, establishing the "residual quality first" system design principle.

(4) An LLM diagnostic report engine enhanced with FMEA knowledge is designed, which fuses CNN classification results, residual evidence, and structured FMEA causal chain knowledge to generate physically interpretable natural language diagnostic reports. FMEA knowledge injection improves causal chain coverage by 105.8%.

## 2. Overall System Architecture

The proposed intelligent fault diagnosis system consists of four core modules, as shown in Fig. 1.

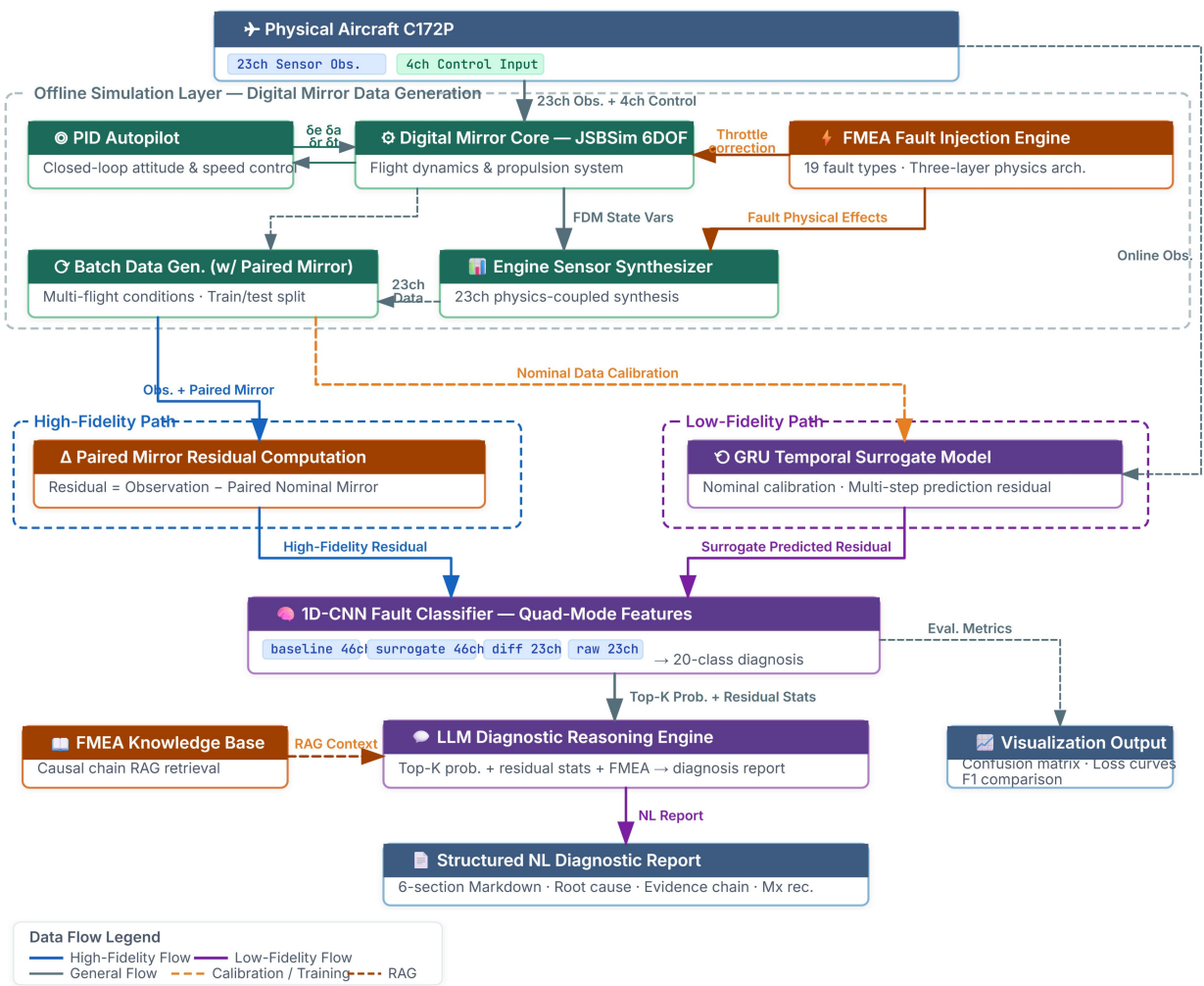

Fig. 1  Overall system architecture

**Digital twin modeling module** (Section 3) is responsible for constructing a high-fidelity flight-engine co-simulation environment. This module centers on the JSBSim 6-DoF flight dynamics engine, integrating a PID autopilot, semi-empirical sensor synthesizer, and FMEA fault injection engine, capable of generating 23-channel engine health monitoring time-series data under diverse flight conditions and fault scenarios.

**Multi-fidelity residual computation module** (Section 4) implements two complementary residual extraction paths. The high-fidelity path uses the digital twin to generate paired nominal mirrors for each observation trajectory, obtaining clean fault residual signals through timestep-wise subtraction. The low-fidelity path uses a GRU temporal surrogate model calibrated on nominal data to predict the normal baseline online, achieving low-latency real-time fault feature extraction through prediction residuals.

**Fault classification module** (Section 5) employs a 1D-CNN network for end-to-end classification of residual feature sequences, achieving automatic identification of 20 fault classes (including nominal).

**LLM diagnostic report engine** (Section 5.2) injects CNN classification results, residual evidence, and FMEA causal chain knowledge into a large language model through structured prompts, generating interpretable diagnostic reports containing physical causal reasoning.

## 3. Digital Twin Modeling

### 3.1. JSBSim flight dynamics and sensor synthesis

The core of the digital twin model is the JSBSim open-source flight dynamics simulation engine, with the Cessna 172P (C172P) as the simulation target. JSBSim implements complete 6-DoF rigid body dynamics, nonlinear aerodynamic models, piston propulsion system models, and standard atmosphere models, capable of high-fidelity simulation of aircraft dynamic responses under various operating conditions. The simulation runs at a time step of 0.02 s, with each simulation lasting 300 s.

To simulate the diversity of real flight conditions, the target cruise altitude for each simulation is randomly sampled within the range of 4500–5500 ft, and the target cruise speed is randomly sampled within 90–110 kts. A closed-loop PID autopilot implements three-channel control for altitude hold, speed hold, and heading hold, ensuring stable flight under various initial conditions.

The sensor synthesizer generates 23-channel engine health monitoring sensor values based on core flight state variables output by JSBSim (RPM, thrust, throttle position, fuel remaining, airspeed, climb rate, altitude, ambient temperature, etc.) through semi-empirical physical approximation formulas, covering the electrical system (4 channels), fuel system (3 channels), oil system (2 channels), engine thermodynamics (9 channels: RPM + 4-cylinder CHT + 4-cylinder EGT), and flight parameters (5 channels). Manufacturing deviations between cylinders are modeled through random offsets, and sensor measurement noise is independently superimposed according to channel characteristics.

As shown in Fig. 2, the core design of the digital twin modeling architecture lies in the physical coupling between the sensor synthesis equations and the fault injection engine: fault effects are not directly superimposed on sensor readings but are instead transmitted to the sensor synthesis equations through physical intermediate variables (cooling efficiency, compression ratio factor, mixture ratio offset, friction factor, etc.), naturally propagating to each channel's sensor readings through physical relationships. This design ensures that cross-channel coupling relationships (e.g., oil temperature rise → viscosity decrease → oil pressure drop) emerge naturally, consistent with the behavior of real physical systems.

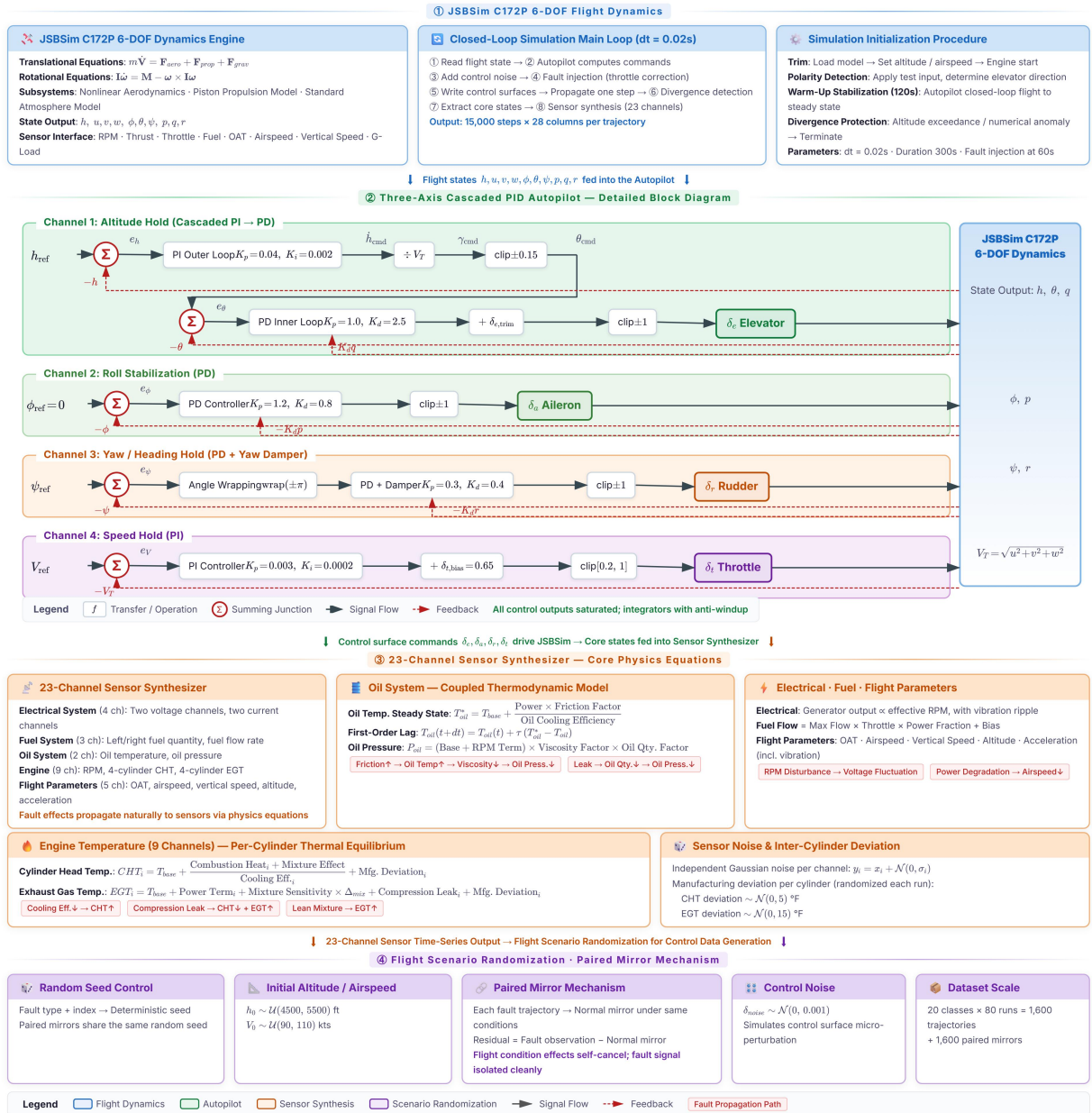

**Fig. 2** Digital twin modeling architecture

## 3.2. FMEA-driven fault injection engine

The fault injection engine is one of the core innovations of the proposed digital twin system, with its design philosophy derived from the failure mode and effects analysis (FMEA) methodology. As shown in Fig. 3, the engine adopts a three-layer architecture:

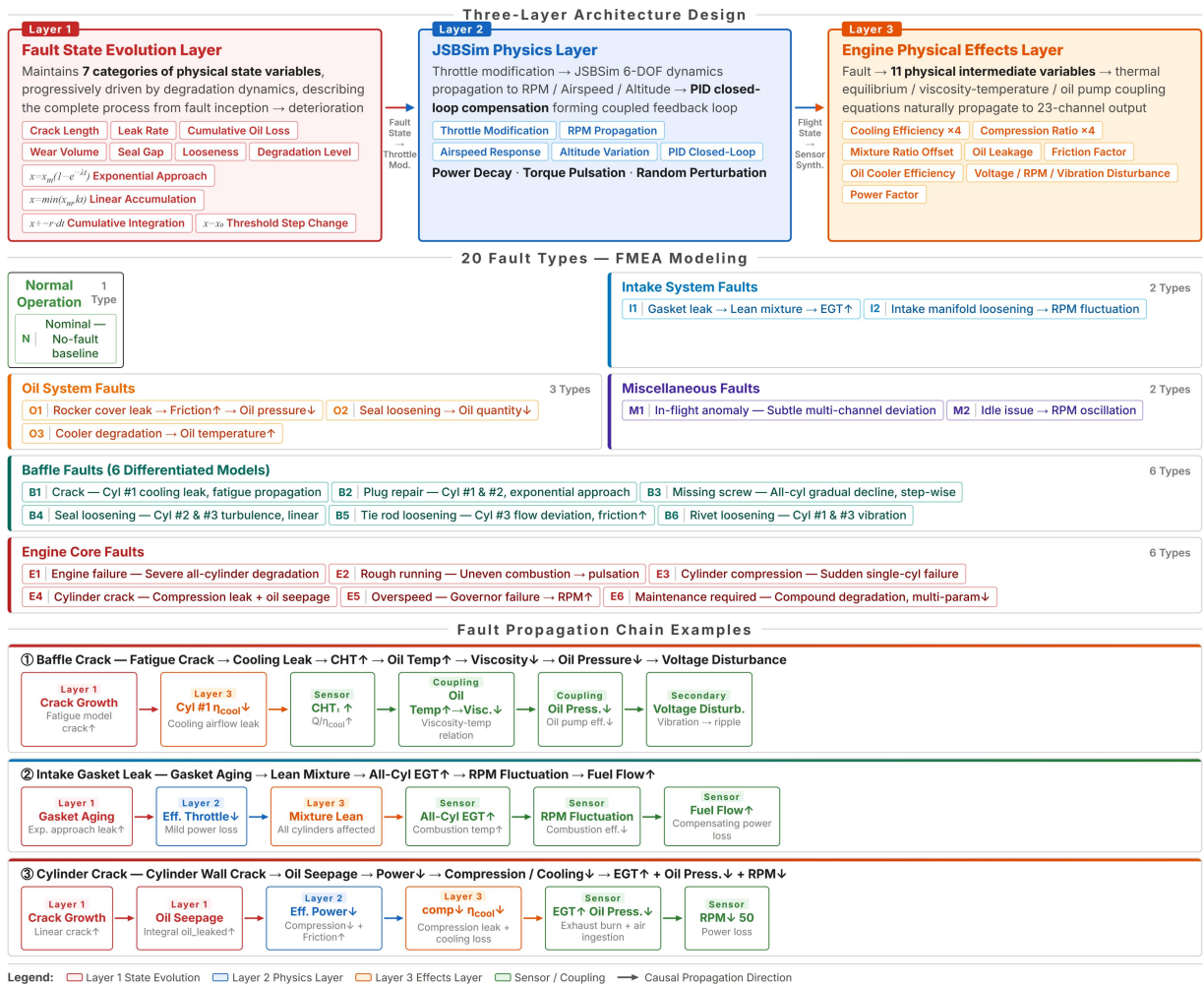

**Fig. 3** Architecture of the FMEA-driven fault injection engine

**Layer 1: Fault state evolution layer.** Each fault type maintains a set of physical state variables (crack length, leakage rate, wear amount, structural looseness, cumulative oil loss, etc.) that evolve over time according to their respective degradation dynamics models. The degradation models include exponential approach (suitable for scenarios with rapid initial deterioration followed by stabilization), linear accumulation (suitable for uniform wear/leakage), fatigue accumulation, and threshold-triggered abrupt change.

**Layer 2: JSBSim physics layer.** Based on the fault state, the impact on the flight control system is computed (primarily the throttle correction coefficient), and power changes are propagated to flight state variables such as RPM, airspeed, and altitude through the JSBSim 6-DoF dynamics engine.

**Layer 3: Engine physical effects layer.** Based on the fault state, a set of physical intermediate variables is computed and transmitted to the sensor synthesizer: per-cylinder cooling efficiency (affecting CHT thermal equilibrium), per-cylinder compression ratio factor (affecting CHT work-generated heat and EGT exhaust combustion), mixture ratio offset (affecting EGT combustion temperature), cumulative oil leakage (affecting the oil volume factor for oil pressure), friction factor (affecting oil temperature and subsequently oil pressure through the viscosity-temperature relationship), oil cooler efficiency (affecting oil temperature heat dissipation), etc. All effects propagate naturally through physical relationships in the sensor synthesis equations, without imposing independent offsets on sensors.

In total, 19 types of engine faults are modeled, covering cooling system faults (6 baffle subcategories), lubrication system faults (rocker cover leakage, engine seal looseness, oil cooler maintenance), intake system faults (intake gasket leakage, intake manifold looseness), compression system faults (cylinder compression issues, cylinder cracks), and comprehensive faults (engine failure, engine running rough, engine overspeed, engine idle problems, in-flight anomalies, engine requires maintenance). Together with the normal operating condition (nominal), there are 20 classes in total.

### 3.3. Paired-mirror data generation mechanism

To achieve high-fidelity residual computation, a paired-mirror data generation mechanism is designed. For each simulation trajectory (whether under fault or normal conditions), the system additionally runs a nominal mirror simulation sharing completely identical initial conditions (target altitude, target speed, random seed). Since the only difference between the two trajectories is whether the fault injection module is activated, their sensor trajectories before fault injection theoretically coincide perfectly, and the residual signal precisely isolates the net impact of fault effects on sensor readings.

For each fault class, 80 simulation trajectories are generated (each lasting 300 s at a sampling rate of 50 Hz), along with 80 corresponding paired nominal mirrors. All data are split into training and test sets at a 70%/30% ratio, with fault trajectories and mirror trajectories sharing the same run_id assigned to the same set.

## 4. Multi-Fidelity Residual Computation Framework

Residual signals serve as the core input features for fault diagnosis in this system. This paper proposes two complementary residual computation paths, high-fidelity and low-fidelity, forming the multi-fidelity residual framework, as shown in Fig. 4.

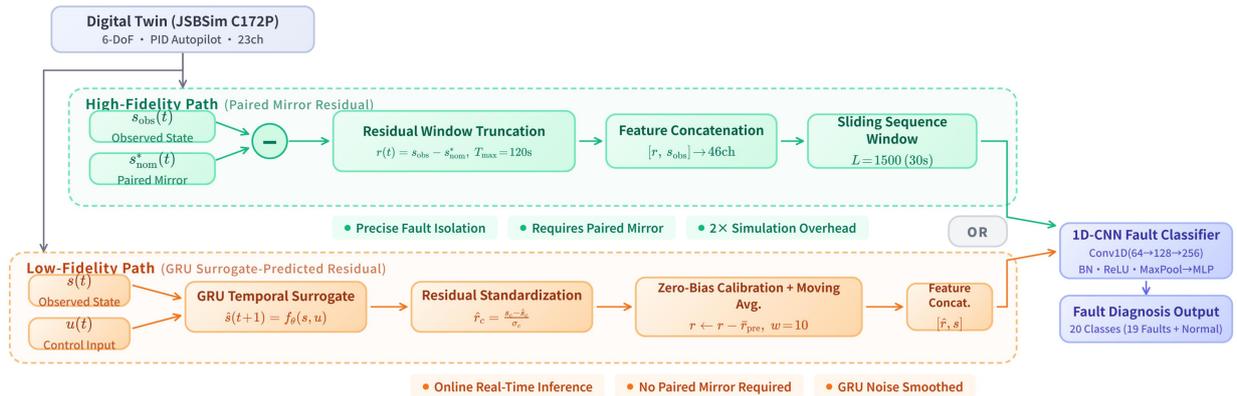

Fig. 4 Schematic of multi-fidelity residual computation

### 4.1. High-fidelity path: Paired-mirror residuals

The paired-mirror residual is defined as the timestep-wise difference between observed sensor values and paired nominal mirror sensor values. Since the observation trajectory and mirror trajectory share completely identical flight dynamics initial states, autopilot parameters, and sensor noise seeds, the mean absolute values across all channels of the paired-mirror residual under nominal conditions are strictly zero. This "zero-bias" baseline implies that any non-zero deviation appearing in the residual necessarily and exclusively originates from the fault effects themselves.

However, fault injection alters the engine thrust characteristics, and the autopilot makes compensatory control responses, causing the flight states of the observation trajectory and mirror trajectory to gradually diverge over time. To address this, the system introduces a residual effective window mechanism, using only residual data within 120 s after fault injection for feature extraction, effectively suppressing the contamination of long-term drift on diagnostic accuracy.

### 4.2. Low-fidelity path: GRU surrogate prediction residuals

The GRU surrogate prediction residual adopts a different technical approach: a GRU temporal surrogate model is calibrated using nominal data from the digital twin, learning the sensor state transition relationships under normal operating conditions, and then the deviation between actual observed values and GRU predicted values is used as the residual signal.

The GRU surrogate model adopts a two-layer GRU architecture (hidden dimension 256) with a fully connected prediction head. The input is a joint sequence of 23-channel sensor states and 4-channel control inputs (27 dimensions), and the output is the predicted 23-channel sensor values for the next H steps. To prevent the model from degenerating into a shallow "copy-last-step" mapping, a multi-step prediction training strategy is employed: the model simultaneously fits sensor values from 1 to H steps ahead during training, forcing it to capture deeper temporal dependencies. During inference, only the 1-step-ahead prediction is used for residual computation.

GRU surrogate residuals inevitably contain model prediction noise. To address this, a three-layer residual normalization mechanism is designed: (1) global standard deviation normalization—calibrating per-channel residual standard deviations on nominal data as fixed baselines; (2) standard deviation lower bound clipping—setting the standard deviation lower bound to 1% of the sensor range to prevent noise amplification in channels with small standard deviations; (3) pre-fault baseline debiasing—using the mean residual before fault injection to eliminate systematic prediction bias. Finally, sliding window mean smoothing is applied to the normalized residuals to suppress prediction noise.

This path offers significant advantages for online deployment—there is no need to run a complete digital mirror simulation for each new trajectory; only GRU forward inference is required to obtain residuals. Both paths share a unified classifier interface (46 channels: 23-channel residuals concatenated with 23-channel raw states), enabling seamless switching based on deployment conditions.

## 5. Fault Classification and Diagnostic Report

### 5.1. 1D-CNN fault classification network

The fault classifier employs a one-dimensional convolutional neural network (1D-CNN) architecture. The network consists of three convolutional blocks, each containing a 1D convolutional layer (kernel size 7, padding 3), batch normalization layer, ReLU activation function, and max pooling layer, with channel numbers of 64, 128, and 256, respectively. The classification head comprises adaptive global average pooling, two fully connected layers (128-dimensional hidden layer), and Dropout (rate 0.15), outputting a 20-dimensional softmax probability vector.

Training employs the Adam optimizer (learning rate $5\times10^{-4}$, weight decay $10^{-4}$), with the learning rate dynamically adjusted by the ReduceLROnPlateau scheduler. An early stopping strategy based on Macro-F1 (patience=40) is adopted, reverting to the best model weights when the validation F1 ceases to improve. The input sequence length is 1500 steps (corresponding to a 30 s observation window).

## 5.2. LLM diagnostic report engine

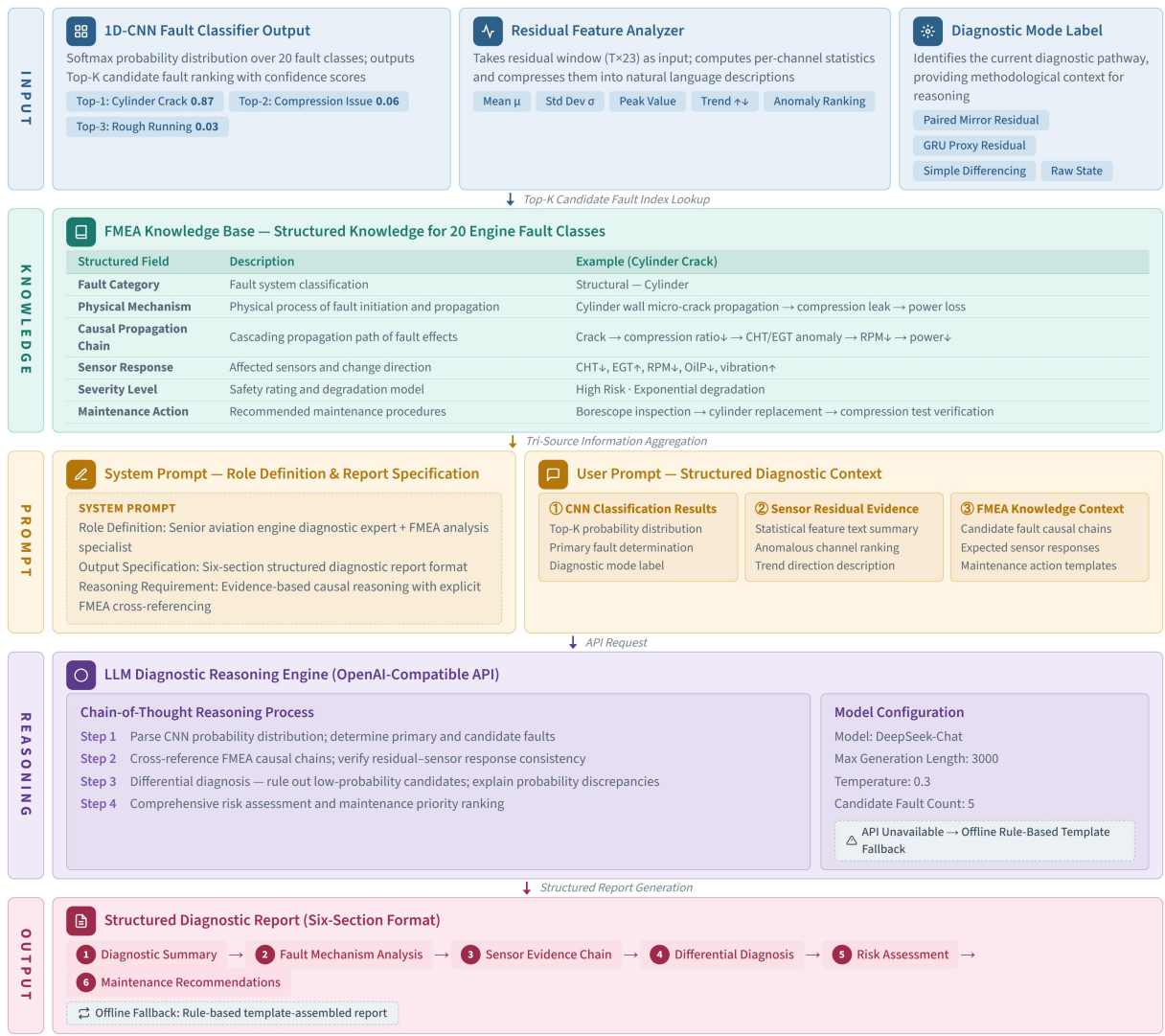

**Fig. 5** Architecture of the LLM diagnostic reasoning and report engine

To enhance the interpretability of the diagnostic system, an LLM diagnostic report engine enhanced with FMEA knowledge is designed. This engine injects the CNN classifier's top-K candidate fault probabilities, key channel anomaly patterns extracted by the residual analyzer, and corresponding fault causal chain knowledge retrieved from the FMEA knowledge base into a large language model (DeepSeek-Chat is used in this work) through structured prompts, generating natural language diagnostic reports containing: (1) primary fault diagnosis conclusion with confidence; (2) physical mechanism analysis based on FMEA causal chains; (3) cross-validation of sensor evidence against causal chains; (4) differential diagnosis (reasoning to exclude other candidate faults); and (5) maintenance recommendations.

The FMEA knowledge base extracts structured information from the domain knowledge of the fault injection engine. Each fault type includes a fault mechanism description, causal propagation chain (complete steps from fault state to sensor response), key affected sensors and their response directions, severity ratings, and maintenance recommendations. This knowledge base serves as the context for retrieval-augmented generation (RAG), enabling the LLM to perform diagnostic reasoning based on physical causal relationships rather than statistical correlations.

## 6. Experimental Results and Analysis

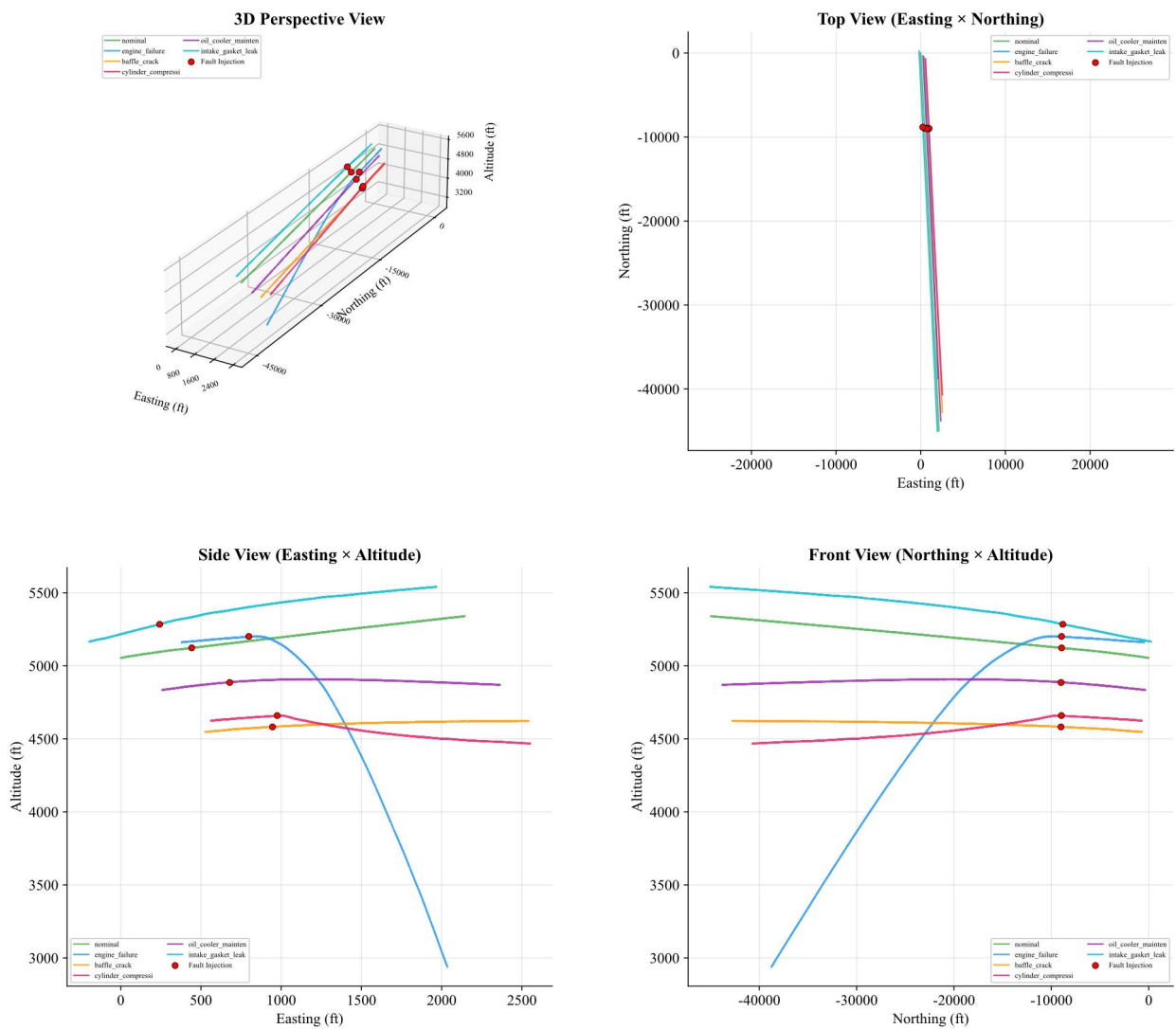

**Fig. 6** Comparison of 3D flight trajectories and side-view altitude profiles under five representative fault types

**6.1. Experimental setup**

The experimental platform uses the JSBSim 1.2.1 flight dynamics engine, PyTorch as the deep learning framework, and Intel Core i7-12700H CPU as the hardware environment. For each fault class, 80 simulation trajectories (including paired nominal mirrors) are generated, split into a training set (1120 trajectories) and test set (480 trajectories) at a 70%/30% ratio, covering 20 fault classes.

**6.2. Physical fidelity validation of simulation data**

To validate the physical plausibility and fault injection fidelity of the digital twin simulation platform, this section systematically analyzes the simulation results of five representative fault types from two perspectives: macroscopic flight trajectory response and microscopic key sensor channel response.

Fig. 6 shows the comparison of 3D flight trajectories between five representative fault types and normal operating conditions. After engine_failure injection, the aircraft altitude decreases continuously and rapidly, exhibiting the most significant altitude loss among all fault types, corresponding to the physical mechanism of exponential decay of effective engine power to near zero. cylinder_compression manifests as a slowly discernible altitude descent trend, consistent with the physical expectation of approximately 7.5% effective power loss. The flight trajectories of baffle_crack and oil_cooler_maintenance nearly coincide with the normal condition, as they primarily affect the cooling/lubrication subsystems rather than thrust output. The intake_gasket_leak trajectory is slightly lower than normal but remains stable.

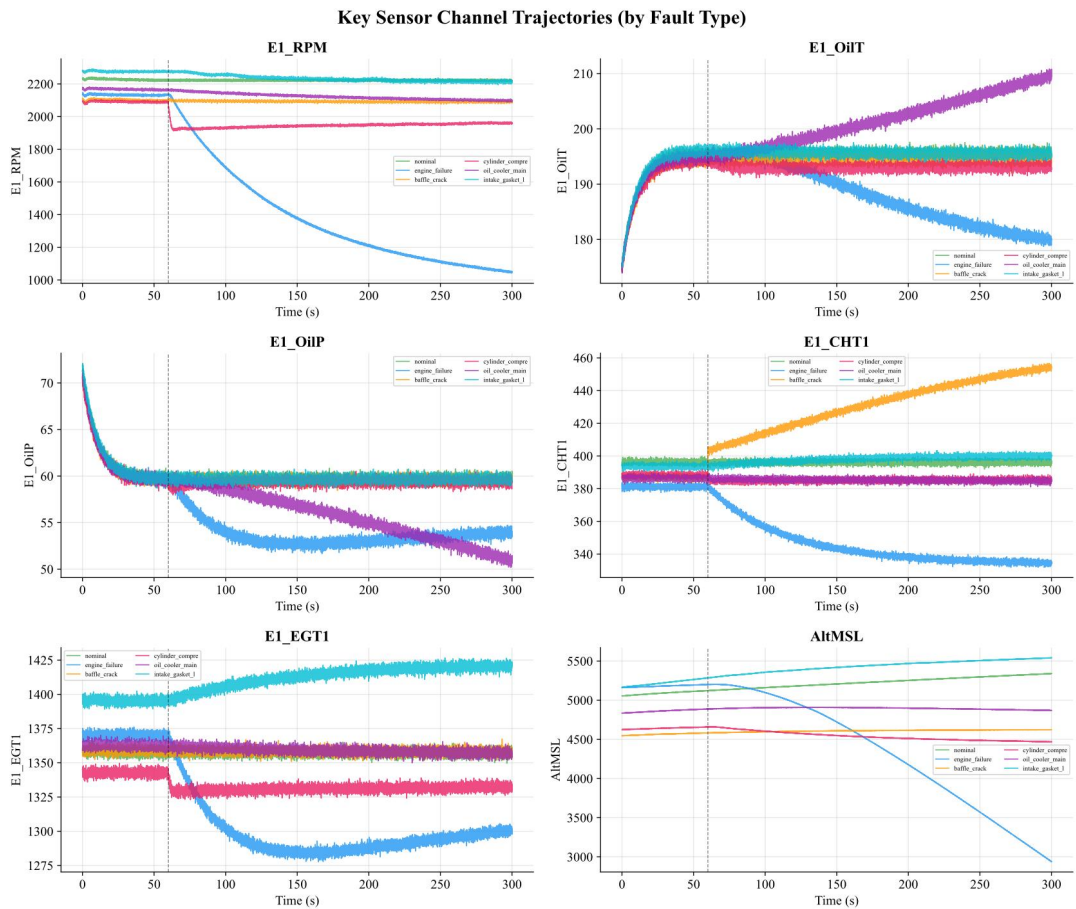

**Fig. 7** Time-series trajectories of six key sensor channels under five representative fault types

Fig. 7 shows the fault response characteristics of six key sensor channels. engine_failure exhibits typical multi-channel coordinated degradation: rapid RPM drop, oil pressure decrease, overall CHT and EGT decline, and continuous altitude loss. baffle_crack is primarily characterized by a significant rise in Cylinder 1 CHT, while oil temperature and oil pressure remain largely unaffected, reflecting the typical characteristics of a localized cooling fault. oil_cooler_maintenance clearly demonstrates the temperature-pressure coupling chain in the oil system: continuous oil temperature rise → viscosity decrease → symmetric oil pressure decline. intake_gasket_leak features a pronounced EGT rise as its most discriminative characteristic, reflecting the direct effect of a lean mixture ratio shift. The above multi-channel coupled response patterns are highly consistent with real fault characteristics reported in aero-engine fault diagnosis literature.

Comprehensive analysis indicates that the proposed fault injection mechanism demonstrates good fidelity in three aspects: causality of fault propagation paths, gradation of fault severity, and physical consistency of multi-channel coupling relationships, providing a credible simulation data foundation for downstream fault diagnosis.

### 6.3. Classification performance comparison of four schemes

To systematically evaluate the diagnostic effectiveness of residual features at different fidelity levels, this section comprehensively compares four schemes: paired-mirror residuals, GRU surrogate prediction residuals, raw states, and simple differencing. All four schemes employ the same 1D-CNN classifier architecture, differing only in input features.

**Table 1** Overall classification performance comparison of four schemes

| Scheme | Accuracy | Macro-P | Macro-R | Macro-F1 |
|---|---|---|---|---|
| Paired-mirror residuals | 0.962 | 0.964 | 0.962 | 0.961 |
| GRU surrogate prediction residuals | 0.955 | 0.957 | 0.955 | 0.955 |
| Raw states | 0.915 | 0.915 | 0.915 | 0.914 |
| Simple differencing | 0.725 | 0.743 | 0.725 | 0.726 |

As shown in Table 1, the performance of the four schemes exhibits a clear hierarchical structure. The paired-mirror residual scheme ranks first with a Macro-F1 of 0.961, closely followed by the GRU surrogate prediction residual at 0.955, with a marginal difference of only 0.6 percentage points. The raw state scheme achieves a Macro-F1 of 0.914, showing approximately 4–5 percentage points of performance degradation compared to the two residual schemes. The simple differencing scheme performs worst (0.726), far below the other three schemes.

The significant advantage of the two residual schemes lies in their effective purification of fault signals through constructing normal baseline references. The paired-mirror residual obtains the cleanest fault deviation signals by directly subtracting nominal mirror trajectories under identical initial conditions. The GRU surrogate residual achieves residual computation by learning sensor state transition mappings under nominal conditions and substituting model predicted values for mirror trajectories. Although this introduces certain model prediction errors, it still maintains diagnostic capability close to the high-fidelity path. The severe degradation of the simple differencing scheme indicates that first-order differencing without physics-based model guidance is essentially a high-pass filter that eliminates slowly evolving fault trend signals along with the DC component.

Confusion matrix analysis indicates that misclassifications of both residual schemes are highly concentrated in the "nominal" class and the "engine requires maintenance" class. The latter has an extremely low degradation rate, producing sensor deviations within the limited observation window that approach noise levels, constituting the most challenging identification target for this diagnostic system. Excluding these two classes, the remaining 18 fault classes of the paired-mirror residual scheme all achieve F1-Scores of 1.000 or close to 1.000.

Notably, the performance degradation from high-fidelity to low-fidelity paths follows a long-tailed distribution: 12 out of 20 classes show no performance loss (F1 change of 0.000), and only 3 classes exhibit losses exceeding 0.01. This indicates that the prediction errors of the GRU surrogate model have noticeable effects only on a few fault types with weak signatures.

### 6.4. Comprehensive comparison with multiple methods

To systematically evaluate the effectiveness of the proposed method, this section conducts comprehensive cross-combination experiments of 4 feature extraction modes and 6 classifier architectures (1D-CNN, LSTM, ResNet, Transformer, SVM, RF) under the same dataset split and evaluation protocol, totaling 24 schemes.

Table 2 Comprehensive comparison results of 24 schemes (ranked by Macro-F1 in descending order, top 12)

| Scheme | Accuracy | Macro-F1 | Per-class F1 std |
|---|---|---|---|
| Paired-mirror residuals + ResNet | 0.967 | 0.967 | 0.099 |
| Paired-mirror residuals + LSTM | 0.965 | 0.964 | 0.107 |
| Paired-mirror residuals + Transformer | 0.964 | 0.963 | 0.107 |
| Paired-mirror residuals + 1D-CNN | 0.962 | 0.962 | 0.111 |
| GRU surrogate residuals + ResNet | 0.958 | 0.958 | 0.103 |
| GRU surrogate residuals + 1D-CNN | 0.958 | 0.957 | 0.122 |
| GRU surrogate residuals + LSTM | 0.952 | 0.951 | 0.130 |
| Raw states + Transformer | 0.952 | 0.951 | 0.121 |
| GRU surrogate residuals + Transformer | 0.949 | 0.949 | 0.149 |
| Paired-mirror residuals + SVM | 0.944 | 0.944 | 0.126 |
| Raw states + ResNet | 0.937 | 0.937 | 0.163 |
| Paired-mirror residuals + RF | 0.918 | 0.919 | 0.139 |

The experimental results present a clear three-tier structure. The first tier comprises deep learning schemes based on residual features (Macro-F1 > 0.95), the second tier includes raw states + deep learning schemes (Macro-F1 approximately 0.87–0.95), and the third tier consists of simple differencing schemes (Macro-F1 < 0.72). Among them, paired-mirror residuals + ResNet ranks first with a Macro-F1 of 0.967.

**Contribution factor analysis.** With the 1D-CNN classifier fixed, switching the feature mode (simple differencing → paired-mirror residuals) causes Macro-F1 to change from 0.716 to 0.962, a variation range of 0.246. With paired-mirror residual features fixed, switching the classifier architecture (RF → ResNet) causes Macro-F1 to change from 0.919 to 0.967, a variation range of only 0.048. The contribution magnitude of the feature mode is approximately 5.1 times that of the classifier architecture. This finding reveals the "residual quality first" system design principle: in engineering practice, prioritizing resources to construct high-fidelity digital twin models for obtaining high-quality residual features yields more significant performance returns than merely pursuing more complex classifier architectures.

It should be noted that the above conclusion holds under the premise that the classifier possesses temporal awareness capability. In the experiments, the "GRU surrogate residuals + RF" scheme (Macro-F1=0.803) performs below "raw states + SVM" (0.870) because Random Forest cannot model cross-channel correlations in the 46-dimensional temporal residual features. Paired-mirror residuals + RF (0.919) does not exhibit similar degradation, indicating that high signal-to-noise ratio residuals have lower dependence on classifier complexity.

### 6.5. Ablation studies

#### 6.5.1. Input feature ablation

This system defaults to using the concatenation of residuals and raw states as classifier input (46 channels). To quantify the diagnostic contribution of residual features, three input configurations are compared under both residual modes, with results shown in Table 3.

Table 3  Input feature ablation experiment results

| Feature configuration | Acc | Prec | Rec | F1 |
| --- | --- | --- | --- | --- |
| Residuals + states (baseline) | 0.959 | 0.960 | 0.959 | 0.959 |
| Residuals only (baseline) | 0.990 | 0.992 | 0.990 | 0.990 |
| States only (baseline) | 0.850 | 0.866 | 0.850 | 0.856 |
| Residuals + states (surrogate) | 0.947 | 0.947 | 0.947 | 0.945 |
| Residuals only (surrogate) | 0.948 | 0.948 | 0.948 | 0.948 |
| States only (surrogate) | 0.862 | 0.864 | 0.862 | 0.862 |

Under the baseline mode, residuals-only input achieves the highest F1 (0.990), significantly outperforming the residuals + states concatenation (0.959) by 3.1 percentage points. This result positively validates the purity advantage of paired-mirror residuals: the residual signal already fully encodes all fault discriminative information across 23 channels, and introducing raw states instead adds redundant dimensions. Under the surrogate mode, the gap between residuals-only and concatenation narrows to 0.3 percentage points, because the prediction noise in GRU surrogate residuals reduces residual purity, and the absolute magnitude information in raw states can provide complementary discriminative evidence.

The core rationale for defaulting to the concatenation configuration lies in the online deployment requirements of the multi-fidelity framework: both paths share the same classifier weights, and the concatenation strategy provides a unified 46-channel input interface, ensuring seamless path switching. When it is determined that paired-mirror residuals are used, the configuration can be switched to residuals-only to unlock the 0.99 performance ceiling.

**6.5.2. GRU multi-step prediction horizon**

Table 4  GRU multi-step prediction horizon ablation experiment results (surrogate mode)

| Prediction horizon H | F1 | MAE | RMSE |
| --- | --- | --- | --- |
| 1 | 0.945 | 1.812 | 2.286 |
| 3 | 0.942 | 1.854 | 2.378 |
| 5 | 0.947 | 1.585 | 2.120 |
| 10 (default) | 0.955 | 1.368 | 1.836 |

The experimental results exhibit two consistent trends: prediction accuracy continuously improves as H increases (MAE decreases from 1.812 to 1.368, a reduction of 24.5%), and downstream classification performance is positively correlated with prediction accuracy (F1 increases from 0.945 to 0.955). The multi-step prediction constraint forces the GRU to learn more robust temporal representations, enhancing the signal-to-noise ratio of surrogate residuals. H=10 achieves optimal performance in both prediction accuracy and classification performance, corresponding to a 0.2 s prediction horizon that covers the characteristic time scale of typical engine thermodynamic responses.

**6.6. Inference speed analysis**

Table 5  Single-sample inference time comparison of four diagnostic schemes (CPU environment)

| Diagnostic scheme | Mirror simulation (ms) | Feature computation (ms) | CNN inference (ms) | Total (ms) |
| --- | --- | --- | --- | --- |
| Paired-mirror residuals | 1885.7 | 5.0 | 2.6 | 1893.4 |
| GRU surrogate prediction residuals | — | 441.4 | 1.1 | 442.6 |
| Simple differencing | — | 4.0 | 1.7 | 5.8 |
| Raw states | — | 3.4 | 1.0 | 4.4 |

The end-to-end inference time of the paired-mirror residual scheme reaches 1893.4 ms, with JSBSim mirror simulation accounting for 99.6% of the computational cost. The GRU surrogate path requires no simulation, achieving an end-to-end time of 442.6 ms and approximately 4.3x acceleration. JSBSim involves sequential iterative CPU-intensive computation that is difficult to parallelize on GPUs, whereas GRU inference is estimated to further accelerate to the 50–100 ms range on GPUs, enabling real-time diagnostic frequencies of 10–20 Hz. This speed advantage makes the GRU surrogate path the preferred choice for online real-time monitoring scenarios, while the paired-mirror path is more suitable for ground-based offline in-depth diagnosis.

**6.7. LLM diagnostic report quality evaluation**

This section evaluates LLM report quality through three comparative experiments: with FMEA knowledge injection (with_fmea), without FMEA knowledge injection (no_fmea), and offline template (offline_template). Evaluation metrics include FMEA causal chain coverage (CCC), sensor evidence consistency (SEC), diagnostic conclusion consistency (DC), and hallucination rate (HR).

FMEA knowledge injection improves causal chain coverage from 0.343 to 0.706 (an increase of 105.8%), demonstrating that the LLM's own parametric knowledge alone cannot systematically cover the specialized causal chains of aero-engine faults. Sensor evidence consistency remains at a high level across all three groups (>0.90), validating the effective guidance of the residual analyzer output format for the LLM. Diagnostic conclusion consistency exceeds 0.94 in all groups, indicating that the LLM faithfully reflects the CNN classifier's output. The hallucination rate decreases by 19.7% after FMEA injection, primarily originating from the LLM citing sensor channels outside the residual analysis results. Case analysis demonstrates that the core added value of the LLM lies in cross-system causal association reasoning and proactive explanation of anomaly absence, capabilities that offline templates do not possess.

Table 6  Summary of automated evaluation metrics for three comparative experiments (mean ± standard deviation)

| Comparison group | CCC | SEC | DC | HR |
| --- | --- | --- | --- | --- |
| with_fmea | 0.706±0.226 | 0.910±0.105 | 0.940±0.068 | 0.122±0.098 |
| no_fmea | 0.343±0.267 | 0.908±0.106 | 0.939±0.100 | 0.152±0.079 |
| offline_template | 0.950±0.218 | 0.973±0.074 | 0.976±0.053 | 0.054±0.074 |

## 7. Conclusions

This paper proposes an intelligent fault diagnosis method for general aviation aircraft based on multi-fidelity digital twin and FMEA knowledge enhancement. The main conclusions are as follows:

(1) Based on the JSBSim flight dynamics engine and an FMEA-driven three-layer fault injection architecture, a digital twin simulation platform covering 19 types of engine faults has been successfully constructed. Fault effects are transmitted to sensor synthesis equations through physical intermediate variables, achieving natural emergence of multi-channel coupling relationships. The qualitative characteristics of the simulation data are highly consistent with real fault characteristics reported in aero-engine fault diagnosis literature.

(2) The multi-fidelity residual framework comprising paired-mirror residuals and GRU surrogate prediction residuals achieves Macro-F1 of 96.1% and 95.5%, respectively, on the 20-class fault classification task. A performance gap of only 0.6% yields 4.3x (CPU) and potentially 19–38x (GPU) inference acceleration, supporting flexible accuracy-speed trade-offs. Ablation experiments demonstrate that using only paired-mirror residuals can reach the performance ceiling of F1=0.99.

(3) Comprehensive comparison experiments across 24 schemes quantitatively reveal that the contribution of residual feature quality to diagnostic performance is approximately 5 times that of classifier architecture selection (feature mode variation range 0.246 vs. classifier architecture variation range 0.048), establishing the "residual quality first" system design principle. With high-quality residual features, performance differences among different deep learning classifiers are compressed to within 0.5%.

(4) The LLM diagnostic report engine enhanced with FMEA knowledge improves causal chain coverage by 105.8%. The generated reports can fuse sensor residual evidence and domain causal knowledge for physically interpretable diagnostic reasoning, providing credible semantic support for diagnostic decision-making in aviation safety-critical domains.

Future work will expand in the following directions: (1) introducing real flight data recorder (FDR) data for statistical benchmarking and parameter calibration of the digital twin simulation model; (2) extending flight envelope coverage (takeoff, climb, descent phases) to improve the representativeness of simulation data; (3) adding inter-cylinder temperature difference feature channels and frequency-domain features to improve the distinguishability of baffle subcategories and oil system faults; and (4) exploring ensemble fusion strategies of the paired-mirror and GRU surrogate residual paths to further reduce the confusion rate for difficult fault pairs.

## References

[1] Wang HF, Jiang W, Deng XY, et al. A new method for fault detection of aero-engine based on isolation forest. Measurement 2021;185:110064.
[2] Liao ZB, Zhan KY, Zhao H, et al. Addressing class-imbalanced learning in real-time aero-engine gas-path fault diagnosis via feature filtering and mapping. Reliability Engineering & System Safety 2024;249:110191.
[3] Duarte D, Marado B, Nogueira J, et al. An overview on how failure analysis contributes to flight safety in the Portuguese Air Force. Engineering Failure Analysis 2016;65:86-101.
[4] Naderi E, Khorasani K. Data-driven fault detection, isolation and estimation of aircraft gas turbine engine actuator and sensors. Mechanical Systems and Signal Processing 2018;100:415-438.
[5] Coutinho PF, Ramos RGF, Ribeiro AMR. Framework for offline data-driven aircraft fault diagnosis. Journal of Aerospace Information Systems 2023;20(3):127-141.
[6] Lu F, Huang JQ, Lv YQ. Gas path health monitoring for a turbofan engine based on a nonlinear filtering approach. Energies 2013;6(1):492-513.
[7] Wang K, Guo YQ, Zhao WL. Gas path fault detection and isolation for aero-engine based on LSTM-DAE approach under multiple-model architecture. Measurement 2023;210:112560.
[8] Hu Y, Miao XW, Si Y, et al. Prognostics and health management: A review from the perspectives of design, development and decision. Reliability Engineering & System Safety 2022;217:108063.
[9] Li YG. Performance-analysis-based gas turbine diagnostics: A review. Proceedings of the Institution of Mechanical Engineers, Part A: Journal of Power and Energy 2002;216(5):363-377.
[10] Lei YG, Yang B, Jiang XW, et al. Applications of machine learning to machine fault diagnosis: A review and roadmap. Mechanical Systems and Signal Processing 2020;138:106587.
[11] Yan S, Zhong X, Shao HD, et al. Digital twin-assisted imbalanced fault diagnosis framework using subdomain adaptive mechanism and margin-aware regularization. Reliability Engineering & System Safety 2023;239:109469.
[12] Zhang YC, Ren ZH, Zhou SH, et al. Supervised contrastive learning-based domain adaptation network for intelligent unsupervised fault diagnosis of rolling bearing. IEEE/ASME Transactions on Mechatronics 2022;27(6):5371-5380.
[13] Xia M, Shao HD, Williams D, et al. Intelligent fault diagnosis of machinery using digital twin-assisted deep transfer learning. Reliability Engineering & System Safety 2021;215:107938.
[14] Huang YF, Tao J, Sun G, et al. A novel digital twin approach based on deep multimodal information fusion for aero-engine fault diagnosis. Energy 2023;270:126894.
[15] Zhang PB, Chen RX, Yang LX, et al. Recent progress in digital twin-driven fault diagnosis of rotating machinery: A comprehensive review. Neurocomputing 2025;625:129467.

[16] Ma C, Zhan XW, Shi HT, et al. Digital twin-assisted enhanced meta-transfer learning for rolling bearing fault diagnosis. Mechanical Systems and Signal Processing 2023;200:110490.

[17] Li CJ, Li SY, Zhang AS, et al. Digital twin-driven partial domain adaptation network for intelligent fault diagnosis of rolling bearing. Reliability Engineering & System Safety 2023;234:109186.

[18] Wang JJ, Ye LK, Gao RX, et al. Digital twin for rotating machinery fault diagnosis in smart manufacturing. International Journal of Production Research 2019;57(12):3920-3934.

[19] Dong YT, Jiang HK, Wu ZH, et al. Digital twin-assisted multiscale residual-self-attention feature fusion network for hypersonic flight vehicle fault diagnosis. Reliability Engineering & System Safety 2023;235:109253.

[20] Huang YF, Tao J, Zhao JY, et al. Graph structure embedded with physical constraints-based information fusion network for interpretable fault diagnosis of aero-engine. Energy 2023;283:129120.

[21] Xia PC, Huang YX, Tao ZY, et al. A digital twin-enhanced semi-supervised framework for motor fault diagnosis based on phase-contrastive current dot pattern. Reliability Engineering & System Safety 2023;235:109224.

[22] Long ZH, Bai MJ, Ren MH, et al. Fault detection and isolation of aeroengine combustion chamber based on unscented Kalman filter method fusing artificial neural network. Energy 2023;272:127117.

[23] Zhou TY, Han T, Droguett EL. Towards trustworthy machine fault diagnosis: A probabilistic Bayesian deep learning framework. Reliability Engineering & System Safety 2022;224:108525.

[24] Liu C, Song J, Tang D, et al. Probing a novel machine tool fault reasoning and maintenance service recommendation approach through data-knowledge empowered LLMs integrated with AR-assisted maintenance guidance. Advanced Engineering Informatics 2025;65:103460.

[25] Guo Z, Wan L, Wang YQ, et al. LMPHM: Fault inference diagnosis based on causal network and large language model-enhanced knowledge graph network. Chinese Journal of Mechanical Engineering 2025;(in press).